%% file: Template.tex
\documentclass{article}
\usepackage{spconf,amsmath,graphicx,hyperref}
\usepackage{lipsum}
\usepackage{amssymb}
\usepackage{booktabs}
\usepackage{xcolor}
\definecolor{darkgreen}{rgb}{0.0, 0.5, 0.0}
\usepackage[table]{xcolor}
\usepackage{makecell}
\usepackage{multirow}
\usepackage{graphicx}
\usepackage{todonotes}
\usepackage{verbatim}

\title{FreeAnimate: Training-Free Human Image Animation with Preview-Guided Denoising}

\name{Yuan Zeng\textsuperscript{1} \qquad Yujia Shi\textsuperscript{2,3} \qquad Zongqing Lu\textsuperscript{1} \qquad QingMin Liao\textsuperscript{1}$^{\dagger}$
\thanks{ $\dagger$ Corresponding author.}}
\address{\textsuperscript{1}Shenzhen International Graduate School, Tsinghua University, China\\
\textsuperscript{2}Harbin Institute of Technology, China\\
\textsuperscript{3}Pengcheng Laboratory, China}

\begin{document}
\ninept
\maketitle
\begin{abstract}
Human Image Animation has seen significant advancements, primarily driven by diffusion models. However, existing methods typically demand substantial training data and resources to achieve high-quality results, limiting generalization and accessibility. In this work, we introduce \emph{FreeAnimate}, a training-free framework that leverages the inherent capabilities of image diffusion models to enable temporal consistency, identity preservation, and background stability. Our approach incorporates a novel preview generation strategy that provides temporal and structural priors from generated preview frames, effectively guiding pose alignment and background consistency without training. Additionally, FreeAnimate introduces Inversion-Boosted Attention and Reference-Anchored Self-Attention modules to guarantee temporal consistency and identity preservation. Experimental results demonstrate that FreeAnimate outperforms existing training-free competitors and training-based baseline methods, achieving generation quality comparable to state-of-the-art methods and offering robust generalization across diverse datasets.  Our project page is at 
\href{https://freeani.github.io/}{https://freeani.github.io/}.

\end{abstract}


\begin{figure}[h] \centering
    \includegraphics[width=0.45\textwidth]{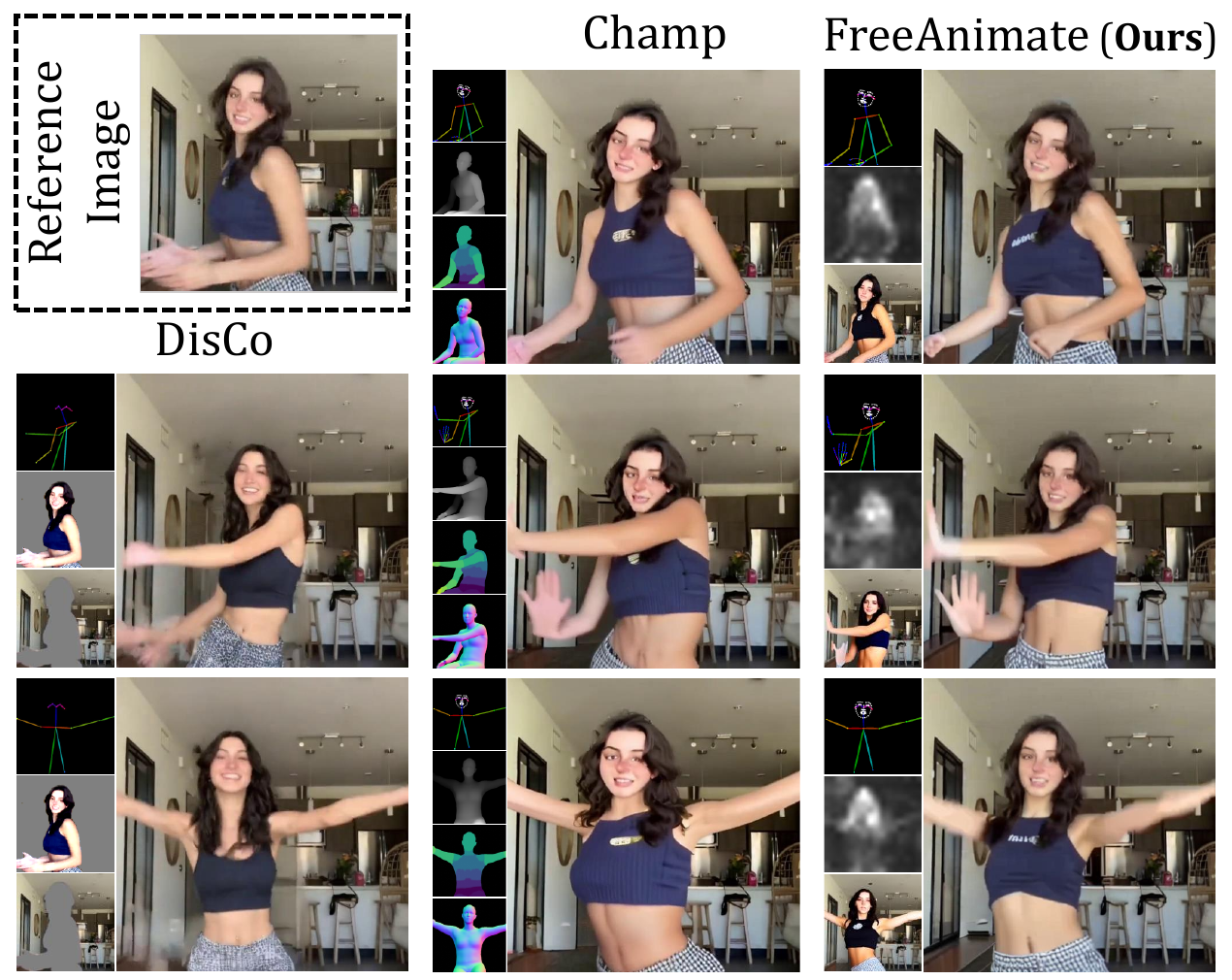}
    \caption{
    FreeAnimate uses preview frames for animation, outperforming DisCo's fixed foreground-background guidance and achieving quality on par with Champ's multi-conditioning generation. Each image's left side shows conditioning inputs,  with the second row of FreeAnimate's inputs representing the cross attention map of preview frames generated by DDIM inversion.
    }
    \label{fig:figure1}
    \vspace{-3mm}
\end{figure}

\section{Introduction}
\label{sec:intro}
The development of diffusion models has greatly advanced applications in content generation, with human-centered themes like Human Image Animation (HIA) receiving particular attention. HIA  \cite{MRAA,TPS,Disco} holds strong potential across various fields, including social media, entertainment industry, and video game development.
Initial attempts with GAN-based HIA \cite{GAN} suffered from unstable training and weak controllability, yielding low-quality results and slowing development of the field.


Recently, diffusion models \cite{DDPM,DDIM} have become the dominant backbone for image and video generation, gradually replacing GANs. With modules such as ControlNet \cite{ControlNet}, they enable strong controllability, and diffusion-based HIA methods like DisCo \cite{Disco} already surpass leading GAN-based approaches \cite{MRAA,TPS}. Building on this, many variants have been proposed: MagicAnimate and MagicPose \cite{Magicanimate,MagicPose} combine U-Net denoising with AppearanceNet and pose ControlNet; AnimateAnyone and MimicMotion \cite{Animateanyone,Mimicmotion} pursue efficiency by replacing heavy ControlNets with lightweight ConvNets; Champ \cite{Champ} integrates 3D parametric cues from SMPL \cite{SMPL}; and StableAnimator \cite{StableAnimator} further enhances identity fidelity with dedicated face encoders and ID adapters.

However, existing methods still rely on large-scale training and resources, and their results often inherit dataset biases. With the rise of powerful foundation models and editing tools, high-quality HIA can now be pursued without extra training. Motivated by this, we propose a training-free framework that leverages pretrained diffusion models. To our knowledge, the only prior training-free approach is PoseAnimate \cite{PoseAnimate}, which ensures pose alignment and temporal coherence through specialized modules. Yet PoseAnimate mainly demonstrates cartoon examples and employs a complex architecture, limiting its applicability to real-world scenarios and further research.

To address these issues, we propose \emph{FreeAnimate}. Unlike prior methods that rely on large datasets or temporal modules \cite{Animatediff}, we introduce a \emph{Preview Generation Strategy} that leverages pretrained models such as Grounded-SAM \cite{GroundedSAM}, editing \cite{Masactrl, T2iAdapter}, and inpainting \cite{Mat} to produce preview frames, providing temporal priors aligned with target frames.
To further improve quality, we design a training-free framework built on an image diffusion model \cite{stablediffusion} with the preview strategy. Unlike methods that add CLIP encoders or AppearanceNet, our framework uses only a pose ControlNet and U-Net for pose-guided synthesis. We also introduce \emph{Inversion-Boosted Attention}, which leverages preview-frame attention maps from DDIM inversion \cite{DDIM} to guide denoising, and \emph{Reference-Anchored Self-Attention}, which preserves identity by letting each latent attend to the reference latent.

As shown in Figure~\ref{fig:figure1}, our approach utilizes inverted noise and attention maps derived from preview frames closest to the target frames, demonstrating superior results over the well-established baseline DisCo and achieving generation quality on par with the state-of-the-art method Champ.
Our contributions are summarized as follows: (1) We propose FreeAnimate, a simple, training-free framework for human image animation. (2) We introduce the Preview Generation Strategy, a framework-agnostic pipeline that generates high-quality preview frames for human image animation, significantly improving both video fidelity and temporal coherence. (3) Experiments on various benchmarks demonstrate FreeAnimate's robust generalization and high-quality generation performance.

\section{Method}
\label{sec:method}
Given a reference image $ {I_{ref}} \in \mathbb{R}^{H \times W \times 3} $ that contains an identity, and a pose sequence ${p_{1:N}} = [p_1, \dots, p_N]  \in \mathbb{R}^{N \times H \times W \times 3} $, where N is the number of frames. Human image animation aims to generate a temporally coherent video ${I_{1:N}} = [I_1, \dots, I_N]$ in which both identity and background appearance follow the reference image, while the identity pose conforms to the given pose sequence.


\subsection{Network Architecture}
\label{sub:NetworkArchitecture}
\textbf{Overview.}
Due to the sampling-based nature of diffusion models, generated images may exhibit unexpected artifacts or deformations. While subtle in single images, these inconsistencies become more noticeable in video. 
To address this issue, many training-based HIA methods \cite{Animateanyone, Champ, Disco, Mimicmotion, StableAnimator} incorporate reference image features into the U-Net via cross-attention, using embeddings from a fixed CLIP image encoder. However, we observe that directly injecting image features without further training yields suboptimal results. 
We attribute this to two primary factors: (1) a domain gap, as the SD model's cross-attention was well-trained with CLIP text embeddings, and (2) semantic-level CLIP image embeddings being too compact to capture fine-grained details.
Some methods \cite{Magicanimate, Champ, FYPv2} use an AppearanceNet, typically a trainable copy of the U-Net, to extract fine-grained reference features. While effective, it adds significant computational overhead.
We propose a simple architecture combining U-Net with ControlNet, which integrates reference image content and structure without CLIP or AppearanceNet. Our method incorporates three components to the basic SD model.

\begin{figure}[tb] \centering
    \includegraphics[width=0.47\textwidth]{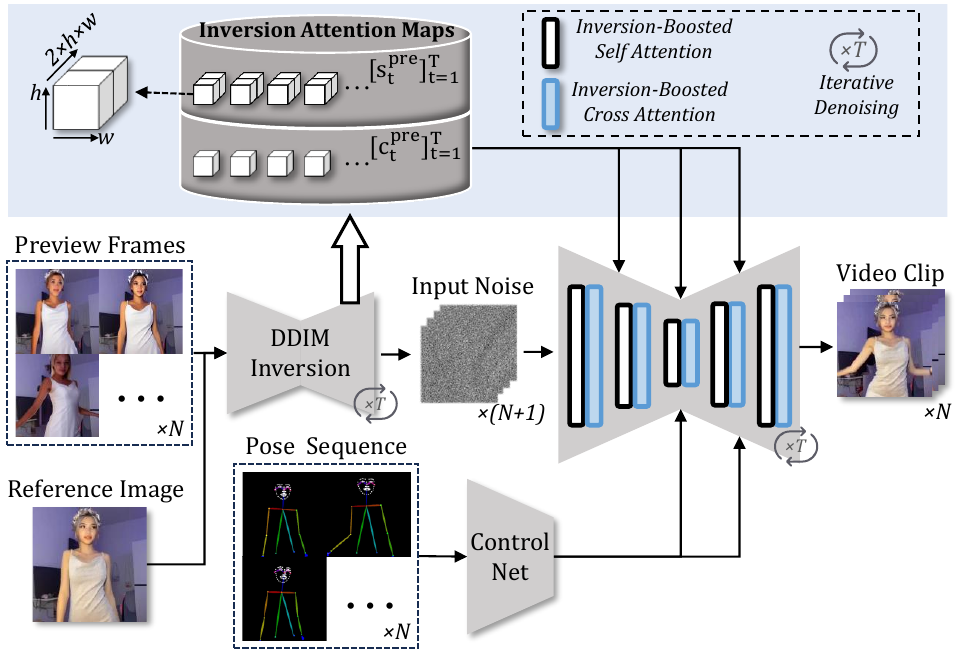}
    \caption{The proposed FreeAnimate framework. Given a reference image and pose sequence, FreeAnimate first generates preview frames using Preview Generation Strategy, then applies DDIM inversion with RA-SA to obtain the initial noise, and stores the resulting attention maps. During denoising, these maps are reused in Inversion-Boosted Attention to preserve structural consistency.} \label{fig:figure2}
    \vspace{-2mm}
\end{figure}

\noindent \textbf{Control Branch.}
In HIA, a key requirement is that generated poses match the input sequence, typically achieved with a control branch. We adopt a well-trained, publicly available ControlNet \cite{ControlNet} to inject pose conditions into the U-Net. Although alternatives like T2I-Adapter \cite{T2iAdapter} or lightweight ConvNets \cite{ControlNetDiscussion188} can offer similar controllability, we directly use ControlNet to ensure a training-free, general, and robust framework.

\noindent \textbf{Inversion-Boosted Attention (IBA).}
Using inverted noise as the initial noise is common in video and image editing. We adopt it from preview frames to enhance temporal consistency and structure. However, feeding it directly into the U-Net causes cumulative DDIM inversion errors that degrade motion and detail \cite{Fatezero}.
To address this issue, we propose Inversion-Boosted Attention (IBA), which enhances structural consistency by modifying both attention mechanisms.
Inspired by FateZero \cite{Fatezero}, IBA offers an effective solution by utilizing DDIM inversion attention maps from preview frames to guide the denoising process. 
As shown in the upper part of Fig.~\ref{fig:figure2}, at each step $t$ of the inversion process, we store the generated self-attention maps $\textit{s}_{t}^{\text{ pre}}$ and cross-attention maps $\textit{c}_{t}^{\text{ pre}}$ from the preview frames $z_{0}^{\text{pre}}$ as follows:
\begin{equation}
z_{T},\left[\textit{s}_{t}^{\text{ pre}}\right]_{t=1}^{T},\left[\textit{c}_{t}^{\text{ pre}}\right]_{t=1}^{T}=\operatorname{DDIM-I\scalebox{0.75}{NV}}\left(z_{0}^{\text{pre}}\right),
\end{equation}
where DDIM-I\scalebox{0.75}{NV} denotes the DDIM inversion process. 
During the denoising process, the inversion attention maps $\left[\textit{s}_{t}^{\text{ pre}}\right]_{t=1}^{T}$ and $\left[\textit{c}_{t}^{\text{ pre}}\right]_{t=1}^{T}$ are used to compute the attention. Formally,
\begin{equation}
\begin{split}
\text{S\scalebox{0.75}{ELF}-A\scalebox{0.75}{TT}}=\operatorname{\scalebox{0.9}{Softmax}} \left(\frac{Q_{s}^{\text{pre}} {K_{s}^{\text{pre}}}^T}{\sqrt{d}}\right) \cdot V_{s} 
=\textit{s}_{t}^{\text{ pre}} \cdot V_{s}, 
\end{split}
\end{equation}
\begin{equation}
\begin{split}
\text{C\scalebox{0.75}{ROSS}-A\scalebox{0.75}{TT}}=\operatorname{\scalebox{0.9}{Softmax}} \left(\frac{Q_{c}^{\text{pre}} {K_{c}^{\text{pre}}}^T}{\sqrt{d}}\right) \cdot V_{c}
=\textit{c}_{t}^{\text{ pre}} \cdot V_{c},
\end{split}
\end{equation}
where \( Q_{s}^{\text{pre}} \), \( K_{s}^{\text{pre}} \), and \( V_{s} \) are the query, key, and value features for self-attention, respectively, and \( Q_{c}^{\text{pre}} \), \( K_{c}^{\text{pre}} \), and \( V_{c} \) are those for cross-attention. \( d \) denotes the feature dimension for attention scaling. 
The terms \( \textit{s}_{t}^{\text{ pre}} \) and \( \textit{c}_{t}^{\text{ pre}} \) represent the precomputed self- and cross-attention maps at time step \( t \), obtained from DDIM inversion over the preview frames. 
Self-attention captures motion and spatial details \cite{Fatezero}, while cross-attention maps provide semantic layouts \cite{Prompt_to_prompt}, reinforcing structure. Through IBA, these signals jointly improve temporal coherence and structural alignment.

\noindent \textbf{Reference-Anchored Self-Attention (RA-SA).}
The control branch and IBA ensure pose alignment and identity preservation, but image diffusion models \cite{stablediffusion} still suffer from temporal inconsistency. To address this, we propose Reference-Anchored Self-Attention (RA-SA), inspired by causal spatial attention \cite{Pix2video,Fatezero}, which enforces frame consistency by anchoring to the reference image.
Specifically, $\text{S\scalebox{0.75}{ELF}-A\scalebox{0.75}{TTENTION}}(Q, K, V)$ for the latent code \( z^{i}_{t} \) of the \( i \)-th frame \((i \in [1, N])\) at denoising time step \( t \in [1, T] \) is implemented in RA-SA as follows:
\begin{equation}
\begin{split}
Q=W^{Q} z^{i}_{t}, K=W^{K}\left[z^{i}_{t} ; z^{a}_{t}\right], V=W^{V}\left[z^{i}_{t} ; z^{a}_{t}\right],
\end{split}
\label{equ:equation4}
\end{equation}
where $W^Q$, $W^K$, and $W^V$ represent the projection matrices from the fixed U-Net, and $\left[\cdot\right]$ denotes the concatenation operation, $z^{i}_{t}$ and $z^{a}_{t}$ denote the latents of current frame and anchor frame. In Pixel2Video \cite{Pix2video} and FateZero \cite{Fatezero}, $a$ is assigned values of $1$ and $\left\lfloor \frac{N}{2} \right\rfloor$, respectively. In our work, to better align with the goal of HIA, we instead designate ${I_{ref}}$ as anchor frame, serving as a consistent reference for the attention mechanism.

It is worth noting that, during the DDIM inversion of preview frames, we also use RA-SA to replace the original self-attention mechanism. This modification results in a change in the dimensions of the self-attention map \( \textit{s}_{t}^{\text{ pre}} \), from $R^{hw \times hw}$ to $R^{hw \times 2hw}$. Due to the combined use of IBA and RA-SA, during the denoising process, both the query and key features in the self-attention mechanism are taken from the DDIM inversion attention maps of the preview frames, while the value features are computed in real-time using the current latent and reference latent, as shown in Eq.\ref{equ:equation4}.

\subsection{Preview Generation Strategy}
\label{sub:PreviewGenerationStrategy}

\begin{figure}[tb] \centering
    \includegraphics[width=0.45\textwidth]{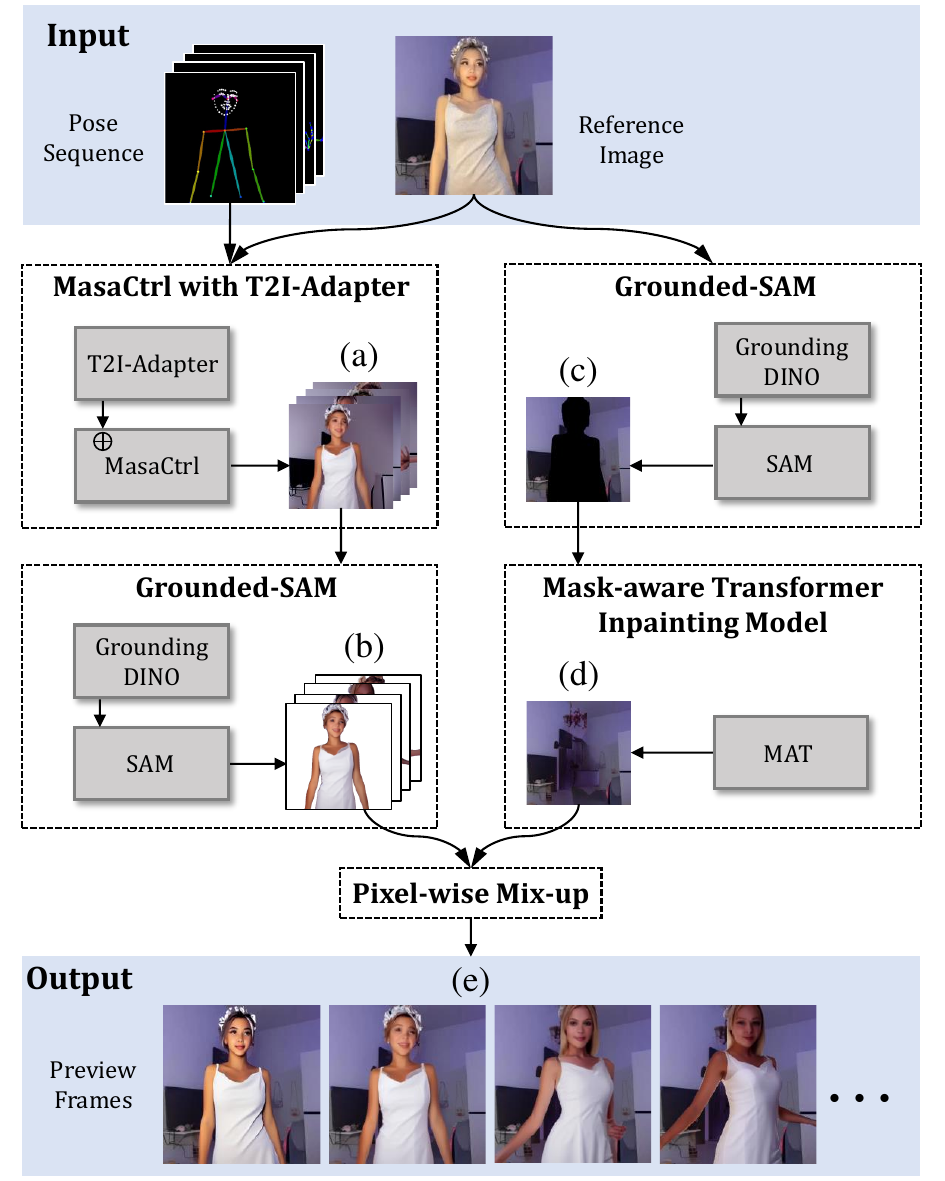}
    \caption{Overview of the Preview Generation Strategy.}
    \vspace{-2mm}
    \label{fig:figure3}
    \vspace{-3mm}
\end{figure}

Initial noise serves as the starting point in the denoising process of diffusion models. Many video editing methods use DDIM inversion to obtain noise for high-quality editing. Since HIA can be viewed as video editing, where each frame is derived from a reference image and pose, using inverted noise from preview frames similar to the target improves fidelity. Motivated by this, we propose the Preview Generation Strategy, a general approach to generate preview frames closely aligned with the target frames, enabling more accurate generation using powerful pretrained models.

The preview generation pipeline is illustrated in Figure~\ref{fig:figure3}. We first use MasaCtrl \cite{Masactrl} with the T2I-Adapter \cite{T2iAdapter} to generate the pose-conditioned image $(a)$. 
The initial noise of this process is obtained from the reference image $I_{ref}$ via DDIM inversion, and the conditioning image comes from the pose sequence $p_i$. In the generated image $(a)$, the appearance matches $I_{ref}$ but the background differs. To address this, Grounded-SAM \cite{GroundedSAM} is used to preserve the foreground, producing image $(b)$, the preview foreground.  
To extract only the background, we first apply Grounded-SAM to remove the identity from $I_{ref}$, producing image $(c)$ with a missing region. The missing part is then filled using the MAT inpainting method \cite{Mat}, generating image $(d)$, the preview background. Although some unrealistic pixels remain due to a domain gap between MAT and our scenario, this does not affect the final result. By mixing the preview foreground and background at the pixel level, the foreground masks imperfect background regions. The final preview frames are obtained as image $(e)$.  
For more preview frame examples, please refer to our project page. 



\begin{table*}[t] \centering
    \newcommand{\Frst}[1]{\textbf{#1}}
    \newcommand{\Scnd}[1]{\underline{#1}}
    \newcommand{\Trd}[1]{\textcolor{darkgreen}{\textbf{#1}}}
    \caption{Quantitative comparison on TikTok and TED-Talks datasets. In the table, $a/b$ denotes results on TikTok and TED-Talks, respectively. \textbf{Bold} text indicates the best result, while \underline{underlined} text indicates the second-best. ↑ indicates that higher values are better. PoseAnimate provides limited experimental details and visual results, with its results (marked in gray) included for reference only. }
    \label{tab:table1}

\input{tables/Table_1}
    \vspace{-1mm}
\end{table*}


\begin{figure*}[tb] \centering
    \includegraphics[width=0.92\textwidth]{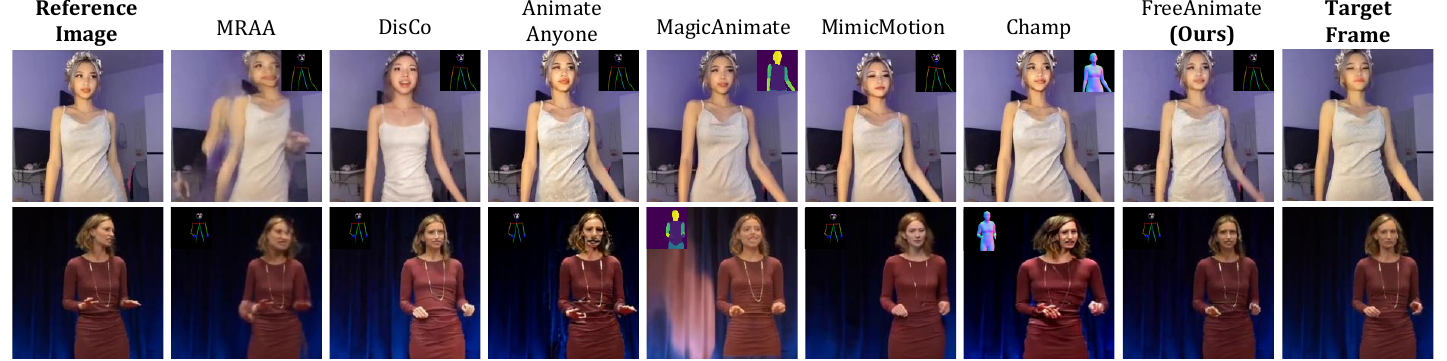}
    \caption{Qualitative comparisons between FreeAnimate and baselines on TikTok (top row) and TED-Talks (bottom row) datasets. The conditioning pose maps are overlaid in the top corner of the generated frames.} \label{fig:figure5}
    \vspace{-5mm}
\end{figure*}

\section{Experiments}
\label{sec:Experiments}

\textbf{Baselines and Benchmarks.}
We chose a diverse set of state-of-the-art HIA methods for comparison.
MRAA \cite{MRAA} and TPS \cite{TPS} are state-of-the-art GAN-based methods. Diffusion-based methods are grouped by training data: (1) DisCo \cite{Disco}, MagicPose \cite{MagicPose}, MagicAnimate \cite{Magicanimate}, and AnimateAnyone \cite{Animateanyone} use only public datasets; (2) Champ \cite{Champ}, MimicMotion \cite{Mimicmotion}, and StableAnimator \cite{StableAnimator} combine public and private data; (3) PoseAnimate \cite{PoseAnimate} is training-free, but since no code or TikTok/TED-Talks visual results are available, we report its results in gray for reference only.
The TikTok dataset \cite{TikTok} contains 350 short dance videos (10 to 15 seconds), mostly with a single subject. For this benchmark, we use the same test set as DisCo \cite{Disco}. The TED-Talks dataset \cite{MRAA} includes a large collection of speaker videos focusing on the upper body, from which we randomly selected 30 clips for evaluation. The EverybodyDanceNow dataset \cite{EverybodyDanceNow} consists of full-body videos of five subjects and is used to assess the generalization of our method to full-body motions.

\noindent
\textbf{Implementation Details.}
We use DWPose \cite{DWpose} for pose extraction and StableAnimator’s \cite{StableAnimator} algorithm for alignment. The first frame is the reference, and others are driving frames, resized to 512 × 512.  
The DDIM sampler is configured with T = 50 steps and a guidance scale of $7.5$. IBA is applied from $t \in[0.5 \times T, T]$, and RA-SA throughout. Experiments are run on an NVIDIA 3090 GPU, with xFormers for memory optimization. More details are available on our project page.

\subsection{Qualitative and Quantitative Results}
\label{sub:all_results}
\textbf{Qualitative Comparison.}
In Figure~\ref{fig:figure5}, we present a qualitative comparison between FreeAnimate and other methods. 
FreeAnimate demonstrates competitive performance against training-based approaches, with visually superior generation quality compared to the well-established baseline, DisCo \cite{Disco}. Since our method is training-free, it exhibits strong generalization across various datasets. Additionally, preview frames provide a strong prior for the background in generated frames, ensuring high alignment between the background of the generated frames and that of the reference image. 

\noindent \textbf{Quantitative Comparison.} 
To reduce bias from erroneous DWPose estimates (e.g., missing hands), we exclude samples with evidently low-quality pose maps. As summarized in Table~\ref{tab:table1}, FreeAnimate is highly competitive despite being training-free. On TikTok dataset, it ranks 2nd or 3rd on FID/SSIM/L1/FVD, outperforming DisCo and MagicAnimate on most image metrics. 
Table~\ref{tab:table1} also presents comparisons on the TED-Talks dataset. 
Since MRAA \cite{MRAA}, TPS \cite{TPS}, and AnimateAnyone \cite{Animateanyone} are trained on the TED-Talks dataset, their performance demonstrates a clear advantage. In contrast, other methods exhibit a noticeable drop on TED-Talks. 
Being training-free, FreeAnimate is less affected by data distribution shifts, maintaining consistent performance across datasets. Built on standard models (SD v1.5, ControlNet v1.1), it does not reach state-of-the-art metrics, but its training-free and modular design enables future gains by replacing components with stronger models.

\subsection{Ablation Study} 

\begin{table}[tb]\centering
    \caption{Ablation study of FreeAnimate.}
    \label{tab:table_ablations}
    \resizebox{0.4\textwidth}{!}{
    \small
    \begin{tabular}{l | l*{5}{c}}
        \toprule
        Method                                      & FID↓  & SSIM↑ & FVD↓ \\
        \midrule
        FreeAnimate \textit{w/o.} preview           & 50.07 & 0.549 & 260.91 \\
        FreeAnimate \textit{w/.} driving            & 23.54 & 0.817 & 147.55 \\
        \rowcolor{gray!15} \textbf{FreeAnimate (Ours)}       & 27.82 & 0.781 & 170.18 \\
        MagicPose \cite{MagicPose}                  & 25.50 & 0.752 & 216.01 \\
        MagicPose \textit{w/.} preview              & 24.61 & 0.760 & 180.49 \\
        \midrule
        \textit{w/o.} Control Branch                & 35.92 & 0.671 & 190.89 \\
        \textit{w/o.} IBA                           & 39.34 & 0.634 & 200.10 \\
        \textit{w/o.} RA-SA                         & 39.20 & 0.602 & 259.02 \\
        \bottomrule
    \end{tabular}
    }
\vspace{-5mm}
\end{table}

All ablation studies are conducted on the TikTok dataset.
The Preview Generation Strategy plays a crucial role in generating high-quality videos by providing initial noise from preview frames, aiding the denoising process with attention maps. Removing this strategy, as shown in Row 1 of Table~\ref{tab:table_ablations}, leads to degraded generation quality due to the use of randomly initialized noise. Using frames from the driving video improves generation significantly (Row 2). This suggests that high-quality preview frames positively impact overall generation. Additionally, applying this strategy to MagicPose results in improved performance compared to the original (Row 5).
Regarding architecture design, FreeAnimate's performance relies on three key components: Control Branch, IBA, and RA-SA. Removing the Control Branch (Row 6) causes a noticeable drop in image quality metrics, though video consistency remains intact. The IBA (Row 7) stabilizes the layout and ensures background consistency, while RA-SA (Row 8) preserves temporal consistency and identity. Removing any of these modules leads to significant performance degradation in both identity preservation and temporal coherence.


FreeAnimate trades higher inference time for zero training cost. We report a stage-wise latency/peak-VRAM breakdown under a standardized setting in Table~\ref{tab:efficiency}. The pipeline is modular: preview/segmentation/inpainting can be replaced by faster alternatives, and diffusion can be accelerated via fewer-step sampling and standard engineering optimizations.

\begin{table}[t]
\centering
\caption{Computational cost analysis.}
\label{tab:efficiency}
\resizebox{0.4\textwidth}{!}{
    \begin{tabular}{l l | c c}
        \hline
        Stage & Key modules & Time/frame & Peak VRAM \\
        \hline
        1 & Preview generation & 3103 ms & 4623 MB \\
        2 & DDIM inversion & 1043 ms & 3017 MB \\
        3 & Denoising & 1350 ms & 5561 MB \\
        \hline
         & All above & 5496 ms & 5561 MB \\
        \hline
    \end{tabular}
}
\vspace{-5mm}
\end{table}

\section{Conclusion}
\label{sec:Conclusion}
We present FreeAnimate, a training-free framework for HIA, addressing limitations associated with data- and resource-intensive HIA methods. By integrating a Preview Generation Strategy with a training-free model architecture, FreeAnimate achieves high fidelity in pose-guided human image animation without relying on extensive training datasets. Through comprehensive evaluations, FreeAnimate consistently outperforms training-free competitors and exhibits generalization across various datasets, illustrating its practical value in diverse applications. 
Future work may focus on enhancing fine detail generation, particularly in facial expressions and hands, and exploring HIA under dynamic and complex backgrounds.

\clearpage

\section{Acknowledgements}
This work is supported by the National Natural Science Foundation of China (U23B2030).

\bibliographystyle{IEEEbib}
\bibliography{refs}

\end{document}

%% file: tables/Table_1.tex
\renewcommand{\arraystretch}{1}
\resizebox{0.9\textwidth}{!}{
\tiny
\setlength{\tabcolsep}{3pt}
\begin{tabular}{l||l||l||*{5}{c}|*{1}{c}}
\toprule
 & Training Data & \textbf{Method} & \textbf{FID↓} & \textbf{SSIM↑} & \textbf{PSNR↑} & \textbf{LPIPS↓} & \textbf{L1↓ (E-04)} & \textbf{FVD↓} \\
\hline
\multirow{2}{*}{GAN} & \multirow{2}{*}{Public}
        & MRAA \cite{MRAA}                      & 85.49 / 50.36 & 0.646 / 0.762 & 28.39 / 31.90 & 0.337 / 0.266 & 4.61 / \Scnd{0.50} & 468.66 / 493.02 \\
        &  & TPS \cite{TPS}                     & 140.37 / 23.71 & 0.560 / 0.771 & 28.17 / \Scnd{32.30} & 0.449 / 0.252 & 6.17 / \Frst{0.49} & 800.77 / 260.67 \\
        \hline
\multirow{8}{*}{SD} & \multirow{4}{*}{Public}
        & DisCo \cite{Disco}                    & 30.75 / 75.48 & 0.668 / 0.575 & 16.55 / 27.99 & 0.292 / 0.309 & 3.78 / 1.2 & 292.80 / 393.04 \\
    &   & MagicPose \cite{MagicPose}            & \Frst{25.50} / \Scnd{23.39} & 0.752 / 0.723 & 29.53 / 30.08 & 0.292 / 0.236 & \Frst{0.81} / 0.81 & 216.01 / 214.23 \\
    &   & MagicAnimate \cite{Magicanimate}      & 32.09 / 41.58 & 0.714 / 0.529 & 29.16 / 28.28 & 0.239 / 0.310 & 3.13 / 1.73 & 179.07 / 223.54 \\
    &   & AnimateAnyone \cite{Animateanyone}    & - / 25.93 & 0.718 / \Frst{0.832} & \Scnd{29.56} / \Frst{33.91} & 0.285 / \Frst{0.159} & - / - & 171.90 / \Frst{80.50} \\
\cline{2-9}
 & \multirow{3}{*}{Public+Private}
        & Champ \cite{Champ}                    & - / 64.96 & \Frst{0.802} / 0.625 & \Frst{29.91} / 21.35 & \Scnd{0.234} / 0.307 & 2.94 / 1.56 & \Scnd{160.82} / 267.03 \\
    &   & MimicMotion \cite{Mimicmotion}        & - / 76.93 & 0.601 / 0.580 & 14.44 / 19.77 & 0.416 / 0.369 & 5.85 / 3.84 & 326.57 / 305.03 \\
    &   & StableAnimator \cite{StableAnimator}  & - / \Frst{22.72} & \Scnd{0.801} / 0.769 & 20.66 / 27.92 & \Frst{0.232} / \Scnd{0.218} & 2.87 / 1.42 & \Frst{140.62} / \Scnd{165.75} \\
\cline{2-9}
 & \multirow{2}{*}{None}
        & PoseAnimate \cite{PoseAnimate}        & \textcolor{gray!60}{31.47 / 21.24} & \textcolor{gray!60}{- / -} & \textcolor{gray!60}{- / -} & \textcolor{gray!60}{- / -} & \textcolor{gray!60}{3.06 / 1.98} & \textcolor{gray!60}{286.33 / 168.02} \\
    &   & \textbf{FreeAnimate (Ours)}           & \Scnd{27.82} / 24.31 & 0.781 / \Scnd{0.803} & 29.26 / 29.18 & 0.244 / 0.231 & \Scnd{2.23} / 1.30 & 170.18 / 169.92 \\
\bottomrule
\end{tabular}
}